# Artificial neural network based modelling approach for municipal solid waste gasification in a fluidized bed reactor

Daya Shankar Pandey[a], Saptarshi Das[b], Indranil Pan[c], James J. Leahy[a], Witold Kwapinski[a,*]

[a]Carbolea Research Group, Chemical and Environmental Science Department, Bernal Institute, University of Limerick, Ireland
[b]Department of Physics, University of Cambridge, JJ Thomson Avenue, Cambridge CB3 0HE, United Kingdom
[c]Department of Earth Science and Engineering, Imperial College London, Exhibition Road, London SW7 2AZ, United Kingdom

*Corresponding author:
Tel.: +353 61234935; fax: +353 61202568.
E-mail address: witold.kwapinski@ul.ie (W. Kwapinski).

**Abstract**

In this paper, multi-layer feed forward neural networks are used to predict the lower heating value of gas (LHV), lower heating value of gasification products including tars and entrained char ($LHV_p$) and syngas yield during gasification of municipal solid waste (MSW) during gasification in a fluidized bed reactor. These artificial neural networks (ANNs) with different architectures are trained using the Levenberg–Marquardt (LM) back-propagation algorithm and a cross validation is also performed to ensure that the results generalise to other unseen datasets. A rigorous study is carried out on optimally choosing the number of hidden layers, number of neurons in the hidden layer and activation function in a network using multiple Monte Carlo runs. Nine input and three output parameters are used to train and test various neural network architectures in both multiple output and single output prediction paradigms using the available experimental datasets. The model selection procedure is carried out to ascertain the best network architecture in terms of predictive accuracy. The simulation results show that the ANN based methodology is a viable alternative which can be used to predict the performance of a fluidized bed gasifier.

1. Introduction:

According to World Bank data, about 4 billion tonnes of waste is generated per year, out of which cities' alone contribute 1.3 billion tonnes of solid waste. This volume is forecast to increase to 2.2 billion tonnes by 2025. Three-fourths of this waste is disposed of in landfills, with only one fourth being recycled. It is expected that in lower income countries waste generation will double in the next 25 years (Hoornweg and Bhada-Tata, 2012). With rapid industrial growth and growing world population, most developing countries are facing acute disposal problem for municipal solid waste (MSW). MSW refers to the discarded materials from household wastes such as kitchen garbage, paper, wood, food waste, cotton as well as materials derived from fossil fuels such as plastic, rubber etc. (Cheng and Hu, 2010). In urban areas significant environmental problems are arising from the disposal of MSW which have led to major concerns regarding human health and environment. These issues are common to both developed as well as developing countries (Pires et al., 2011). Furthermore, these issues are stimulating the need for further development of treatment technologies to meet these global challenges. The new European sustainable development strategy (EU, 2009) promotes thermal treatment processes to recover energy from MSW while tackling the issues related to climate change.

There are several processes that could treat MSW including thermal, biochemical and mechanical processes. Incineration technology is widely used to process MSW, but the control of NOx, SOx, nano-particle, dioxins and furans emissions are challenging (Cheng and Hu, 2010). In a quest for a sustainable waste treatment technology, waste to energy (WtE) technology has been reviewed by (Brunner and Rechberger, 2015). The study concluded that due to the advancement in combustion and air pollution control technologies WtE plants are useful for energy and material recovery from waste without having adverse effects on environment. The impact on the environment of thermal treatment of waste with energy recovery was evaluated by Pavlas et al. (2010) who concluded that thermal treatment of MSW with energy recovery was undoubtedly one of the best techniques. WtE not only offers an alternative to treat the waste but also produces clean energy which can offset primary energy consumption in conventional heat and power units. In general, WtE plants are considered as carbon neutral but they are not. The total carbon content present in the MSW is bound with various materials present in the waste. It was found that more than half of the carbon present is biogenic in nature but the remaining part originates from fossil fuels which cannot be considered as biogenic carbon (Gohlke, 2009). As per the EU's new directive, each WtE plant has to report how much electricity was produced from the renewable sources present in the waste feed. The measured biogenic $CO_2$ fraction in the flue gas from an incinerator plant in Netherlands was between 48-50% (Palstra and Meijer, 2010) whereas, in Austria the ratio of biogenic to anthropogenic energy content in MSW was reported in the range 36-53% (Fellner et al., 2007).

Thermal treatment technologies for MSW have been extensively reviewed by Arena (2012); Leckner (2015); Lombardi et al. (2015); Malkow (2004) and it was proposed that an alternative to combustion is to gasify the MSW for energy recovery. To date, gasification processes have been investigated by several contemporary researchers and extensively reviewed by Gómez-Barea and Leckner (2010). Thermal gasification provides flexibility for the production of heat and power based on clean biomass derived syngas (Basu, 2010). In addition, thermochemical conversion technologies can reduce the original volume of wastes disposed by 80-95% along with energy recovery (Rand et al., 1999). Lately, gasification of solid wastes which originates from the household or industrial sectors have received increasing attention by researchers. The syngas from MSW can be used for heating and production of electricity to offset the use of fossil fuels. However, gasification of MSW is not widespread. The major barrier that has prevented the widespread uptake of advanced gasification technologies for treating MSW has been the higher ash content in the feed making the gasification operation difficult. In addition, high amounts of tar and char contaminants in the produced gas make it unsuitable for power production using energy efficient gas engines or turbines.

A comprehensive review of fluidized bed biomass gasification model was presented by Gómez-Barea and Leckner (2010). In the past, different modelling approaches starting from black box modelling to thermodynamic equilibrium, kinetic rate, fluid-dynamics, neural network and genetic programming models (Pandey et al., 2015; Puig-Arnavat et al., 2010) and Gaussian process based Bayesian inference (Pan and Pandey, 2016) were applied for modelling gasification. These models were validated using pilot scale gasification data. Simulating MSW gasification is computationally expensive and fast meta-models are required. In this paper an artificial intelligence technique namely feedforward neural network is used to predict the heating value of gas (LHV), heating value of gasification products ($LHV_p$) as well as the syngas (product gas) yield. $LHV_p$ is defined as the sum of the LHV of gas and the calorific value of unreacted char (entrained) and tar.

ANN models are not based on modelling the physical combustion and transport equations governing the reactor but they are a class of generic nonlinear regression models which learns the arbitrary mapping from the input data on to the output to obtain computational models with high predictive accuracy. Although ANN based models have been extensively used in other scientific fields, it has only recently gained popularity in renewable energy related applications (Kalogirou, 2001). ANN based models were developed for predicting the product yield and gas composition in an atmospheric steam blown biomass fluidized bed gasifier (Guo et al., 2001). It was concluded that the feed forward neural network (FFNN) model has better predictive accuracy over the traditional regression models. An FFNN model was employed to predict the lower heating value of MSW based on its chemical composition (Dong et al., 2003). ANN was applied for predicting the gasification characteristics of MSW (Xiao et al., 2009) and tested for its feasibility. ANN methodology was used to predict future MSW quality and composition in Serbia to achieve the targets for waste management set by national policy and EU directive by 2016 (Batinic et al., 2011). Two different types of ANN based data-driven models have been developed for the prediction of gas production rate and heating value of gas in coal gasifiers (Chavan et al., 2012). Recently, ANN based predictive tools have been used in fluidized bed gasifiers to predict the syngas composition and gas yield (Puig-Arnavat et al., 2013). The ANN technique has been applied in the gasification area and has shown better results compared to the conventional process modelling approaches. A brief overview of different modelling approaches and their pros and cons is presented in Table 1.

Table 1. Pros and cons of different gasification modelling approach (Gómez-Barea and Leckner, 2010; Robert et al., 2014)

| Modelling approaches | Advantages | Disadvantages | Models using this approach |
|---|---|---|---|
| Black Box model | Independent of gasifier type. Easy to implement. Fast convergence. Widely used for the gas prediction and heating value. | Only applicable for stationary process. Does not provide insight into the gasification process. | Equilibrium model, Thermodynamic model, Pseudo-equilibrium model |
| Kinetic model | Realistic model, which can be used for process design | Depend on reaction kinetics and gasifier | Uniform conversion model |

| | and scaling-up. | type. | Shrinking core model etc. |
|---|---|---|---|
| Fluidisation model | Offers a trade-off between precision and numerical complications. | Applicability of the correlations used has limited scope. | Davidson–Harrison model, Kunii–Levenspiel model etc. |
| Computational fluid-dynamics (CFD) model | Useful in improving the details of the gasifier. | Computationally expensive, time consuming and uncertainty involved with the parameters in closure. | Direct numerical simulation, Large eddy simulation, Two fluid model, Euler-Euler model, Euler-Lagrange model etc. |
| ANN model | Do not need extensive understanding of the process. High predictive accuracy. | Dependent on quantity of datasets. No proper physical interpretation of models can be made. | Feed-forward neural network, Hybrid neural network etc. |

Most of the mathematical models for fluidised bed gasifier are based on the law of conservation (mass, energy and momentum) and other boundary conditions (Gómez-Barea and Leckner, 2010). Depending on the complexity, the model can be a 3-D fluid dynamic model or kinetic rate based model or less complex such as an equilibrium based model. Due to the inherent complexity of gasification processes, development of mathematical models are still at a nascent stage. The aim of this research is to develop neural network based models which can be used to simulate the gasification process with improved accuracy. In this study, computational models derived from artificial intelligence techniques are exploited to learn the nonlinear mapping problem. These types of models can predict the performance of complex systems (including gasification). Therefore, this study is focused on exploiting the potential of the ANN technique to estimate the performance of MSW gasification in a fluidized bed reactor.

**2. Material and methods**

**2.1 Theory of artificial neural network based modelling approach**

ANN is a biologically inspired computational technique that imitates the behaviour and learning process of the human brain. ANNs are universal approximators and their predictions are based on prior available data. It is therefore preferred in many data driven research applications over other theoretical and empirical models where predictive accuracy is of prime concern. The ANN technique has been extensively used in several applications in the fields of pattern recognition, signal processing, function approximation, weather prediction and process simulations (Guo et al., 1997). The recent developments and potential application of ANN in diverse disciplines has motivated the present study. However, application of the ANN technique for modelling of MSW gasification is rarely reported in the literature. ANNs are essentially supervised learning methods, i.e. given an input and an output dataset; they have enough flexibility to model the nonlinear input output mapping. The methodology is generic and does not have any limitation to the type of dataset or the number of input-output variables. These generic ANN models provide flexibility to include other process parameters like tars, unconverted carbon, steam-to-biomass ratio (in the case of steam gasification) etc. or any other process parameter which are deemed necessary (Puig-Arnavat et al., 2013). However the models might not work well for a drastically new configuration of gasifier which is not similar to the training dataset. Nevertheless, this is a limitation of the dataset and not of the ANN based modelling methodology.

Figure 1 represents the multilayer feed-forward neural network architecture with multiple input and multiple output (MIMO) variables. For multiple input and single output (MISO) models the number of output is set to one. It consists of an input layer, multiple hidden layers and an output layer. Each node (neuron) other than the input nodes are equipped with a nonlinear transfer function. Neurons $x_i$ in the input layer distributes the input signals to neurons in the hidden layer $(j)$, while neurons in hidden layers sum up its input signal ($x_i$) after multiplying them by their weight $w_{ij}$. The output $(y_j)$ of the ANN model can be represented as follows (1).

$$y_j = f\left( \sum_{i=0}^{d^{(l-1)}} \left( w_{ij}^l x_i^{l-1} \right) \right) \tag{1}$$

where $f$ is a simple threshold function which can be a sigmoid, hyperbolic tangent or radial basis function, $d$ is the dimension of the network, $l$ represents the number of layers and $w_{ij}^l$ is the weight which belongs to network with $l$ layer and having $i$ input and $j$ hidden layers. The mathematical representation of the ANN model weights can be depicted as (2).

$$w_{ij}^l \in \begin{cases} 1 \leq l \leq L & layers \\ 0 \leq i \leq d^{l-1} & input \\ 1 \leq j \leq d^l & output \end{cases} \tag{2}$$

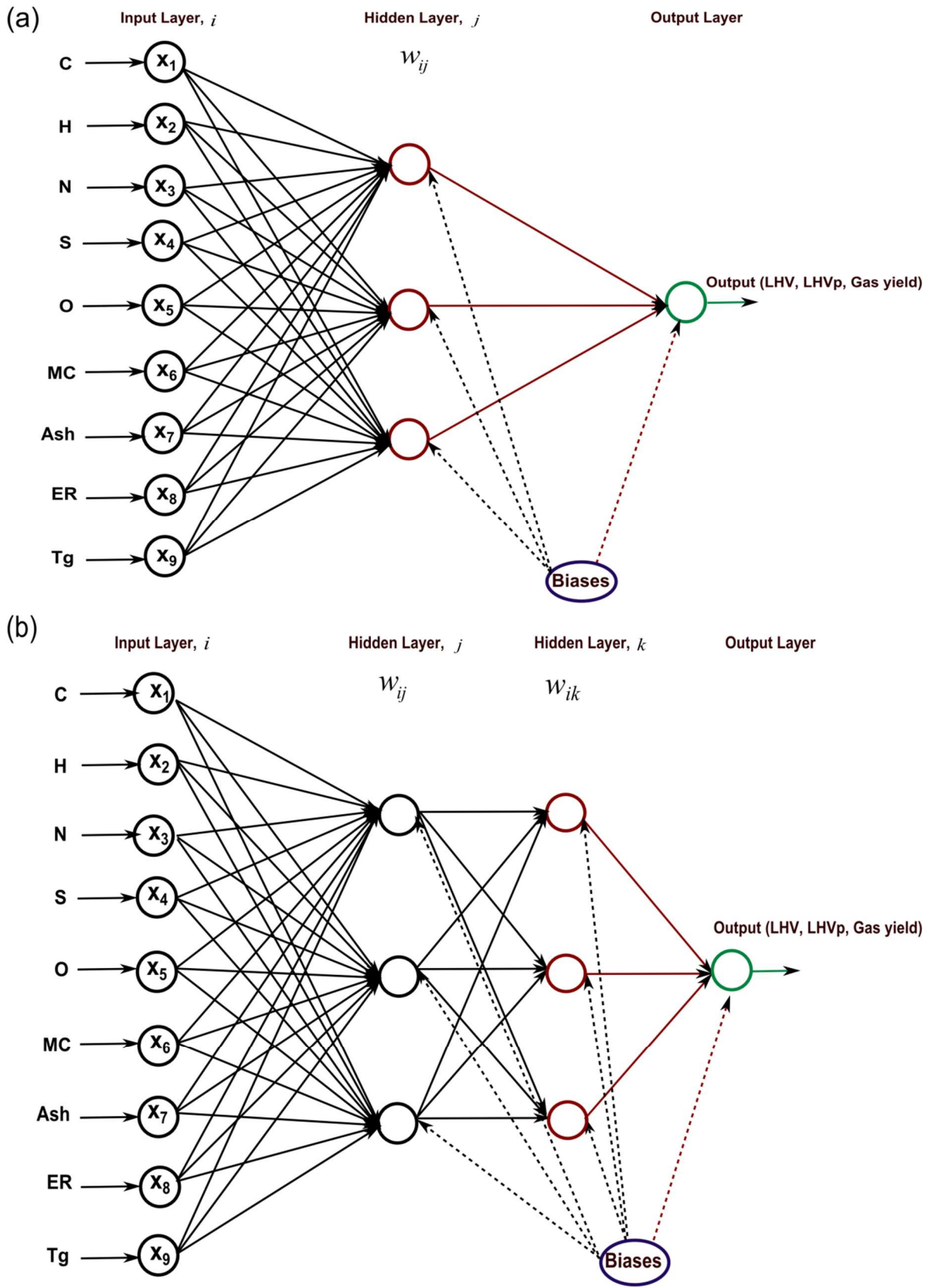

Figure 1. Schematic diagram of the MIMO ANN model (a) single hidden layer (b) double hidden layer.

The input and validated datasets were obtained from a lab-scale fluidized bed gasifier. Those experiments were performed in a lab-scale fluidized bed gasifier (560 mm high with an internal diameter of 31 mm) operating at atmospheric pressure. Heat was supplied from an external source (electric heater) to maintain the temperature of the gasifier. Silica sand was used as a bed material (particle size 0.250–0.355 mm). The gasifier consists of an electric heater, screw feeder to supply the feed, filter for collecting elutriated char and ash and gas-bag for off-line sampling of produced gas. The reported product gas yield was estimated by $N_2$ balance. The details of the gasifier can be found elsewhere (Xiao et al., 2009). Hong Kong MSW data was extracted from (Choy et al., 2004) where MSW was gasified in a small scale gasifier to assess the feasibility of installing an MSW gasifier in Hong Kong University of Science and Technology. Experiments were performed at different temperatures (400 ≤ temperature ≤ 800 °C) and equivalence ratios (0.2 ≤ ER ≤ 0.6).

The modelling methodology using ANN is divided into a training phase and a validation phase. For checking the accuracy and generalization capability of the model, the experimental dataset is divided into training (70%) with the remainder for the validation (15%) and testing (15%) purposes. The input $(x_i)$ and output $(y_i)$ parameters are normalized with respect to the maximum value, to ensure that all data used for training of the network lie within a range of $[0,1]$. The datasets used for the training, testing and validation purpose of the model are randomised. A hyperbolic tangent sigmoid function (*tansig*) $f(x) = \dfrac{e^x - e^{-x}}{e^x + e^{-x}}$ and logarithmic sigmoid function (*logsig*) $f(x) = \dfrac{1}{1 + e^{-x}}$ are used in the hidden layers whereas a pure linear function (*purelin*) is used in the output layer. Both the *tansig* and *logsig* transfer functions are traditionally used and make the ANNs a universal function approximator given a sufficient number of hidden nodes. However, depending on the nature of the data, amongst these two transfer functions, one may outperform the other. Therefore, both the transfer functions are exploited in finding the best suited one for fitting this data.

There have been exhaustive studies on using different training algorithms for ANNs, e.g. Levenberg-Marquardt (LM), scaled conjugate gradient (SCG), Broyden-Fletcher-Goldfarb-Shanno quasi-Newton (BFGS), gradient descent with momentum and adaptive learning rate (GDX), amongst many others (Plumb et al., 2005). The LM gives accurate training results for moderate size neural networks. The other algorithms have disadvantage of slower convergence speed particularly for large networks. In the LM, the Jacobian ($J$) is calculated using the backpropagation technique described in (Hagan and Menhaj, 1994) followed by Hessian ($H = J^T J$) and gradient ($g = J^T e$) calculation, $e$ being the network error. The network weight and bias terms ($x$) are then updated as (3):

$$x_{k+1} = x_k - \left[ J^T J + \mu I \right]^{-1} J^T e \tag{3}$$

where, $\mu$ is a scalar, whose zero or large values make the training algorithm similar to Newton's method, using approximate Hessian or gradient descent with small step size respectively. After each successful step the value of $\mu$ is decreased or alternatively increased if the cost function is not decreased in a step. Based on the above reason, the LM back-propagation training algorithm is used here for minimising the mean squared error (MSE) between the network output and target output.

To develop the ANN model, the nine process parameters that have been used as model inputs are carbon ($x_1$, wt%), hydrogen ($x_2$, wt%), nitrogen ($x_3$, wt%), sulphur ($x_4$, wt%), oxygen ($x_5$, wt%), moisture content ($x_6$, wt%), ash ($x_7$, wt%), equivalence ratio ($x_8$, ER) and the temperature of the gasifier ($x_9$, $T_g$ $^0$C). ER is defined as the ratio between the actual air fed to the gasifier and the air necessary for stoichiometric combustion of the biomass. The input parameters are represented as an input vector $\hat{x}_i = [x_1, x_2, x_3, x_4, x_5, x_6, x_7, x_8, x_9]$ and the output variables are LHV of product gas ($y_1$ kJ/Nm³), LHV$_p$ ($y_2$ kJ/Nm³) and gas yield ($y_3$ Nm³/kg). The input and output variables are in different units. The mean and standard deviation values provide the statistic summary of the dataset to facilitate the reproducibility. The statistical analysis of the input $(\hat{x}_i)$ and output variables $(y_1, y_2, y_3)$ are represented by the mean vectors $\mu_x$ and $\mu_y$, respectively and are given in equations (4) and (5).

$$\mu_x = [43.815 \quad 5.11 \quad 0.685 \quad 0.17 \quad 36.53 \quad 4.21 \quad 9.55 \quad 0.4 \quad 581] \tag{4}$$

$$\mu_y = [3153 \quad 7273 \quad 2.86] \tag{5}$$

Similarly, their corresponding standard deviations are given by $\sigma_x$ and $\sigma_y$ in equations (6) and (7).

$$\sigma_x = [0.1202 \quad 0.6929 \quad 0.5868 \quad 0.1838 \quad 6.2649 \quad 5.9538 \quad 10.6773 \quad 0.2828 \quad 154.1493] \tag{6}$$

$$\sigma_y = [835.80 \quad 4556.60 \quad 2.62] \tag{7}$$

### 2.2 Proposed approach of ANN based learning methodology and optimisation of the model parameters

The MISO and MIMO configurations are used for training of multilayer neural network models. In the MISO case, 9 inputs and 1 output are modelled. Therefore 3 separate ANNs are trained for each of the three cases of LHV, LHV$_p$ and syngas yield. For the MIMO case, the network is trained with 9 inputs and 3 outputs. Therefore, one single network is capable of predicting all three outputs.

Different sets of internal network parameters have been used while training the ANN model *viz.* number of hidden layer, number of neurons in the hidden layer and transfer function, learning rate etc. Deciding the number of neurons in the hidden layer is an important issue in the selection of the neural network architecture and their choice varies on a case by case basis. A detailed study of the effect of internal parameters on the performance of back propagation networks (Hornik et al., 1989) and the procedure involved in selecting the best network topology has been described elsewhere (Maier and Dandy, 1998). The network architecture has a huge influence on the trade-off between predictive accuracy on the training dataset and generalisation capability of the model on untrained data. Hence, both the number of hidden layers and number of neurons in each of these hidden layers must be carefully considered. Having too few neurons in the hidden layer can give rise to lower predictive accuracy (i.e. the network cannot capture the nonlinear trends in the dataset), on the other hand too many neurons in the hidden layers can also result in problems. A highly complex model can suffer from over-fitting the training dataset and it takes much more computational time to train large networks. Hence, a trade-off needs to be found in order to determine the numbers of layers, number of neurons in each layer and transfer function used in the hidden layer. In the past, a trial and error method was employed by other researchers to decide the number of neurons and hidden layers but selection of the optimum layer/hidden node combination was not clear (Azadi and Karimi-Jashni, 2016; Azadi and Sepaskhah, 2012; Puig-Arnavat et al., 2013). An increase or decrease in number of neurons in the hidden layer using the trial and error method cannot accurately identify the best bias-variance trade-off architecture of the ANN. A different approach with a rigorous cross validated accuracy check can be employed while sweeping the number of hidden nodes in single and double layer configuration selecting the best representative model first. This increases the computational load, as reported in this paper, due to the aim of finding out the best ANN architecture to best capture the underlying patterns of this data. This is intrinsically different from what already exists in the literature and also advances the traditional supervised learning data analysis workflow, where the right model is not precisely known.

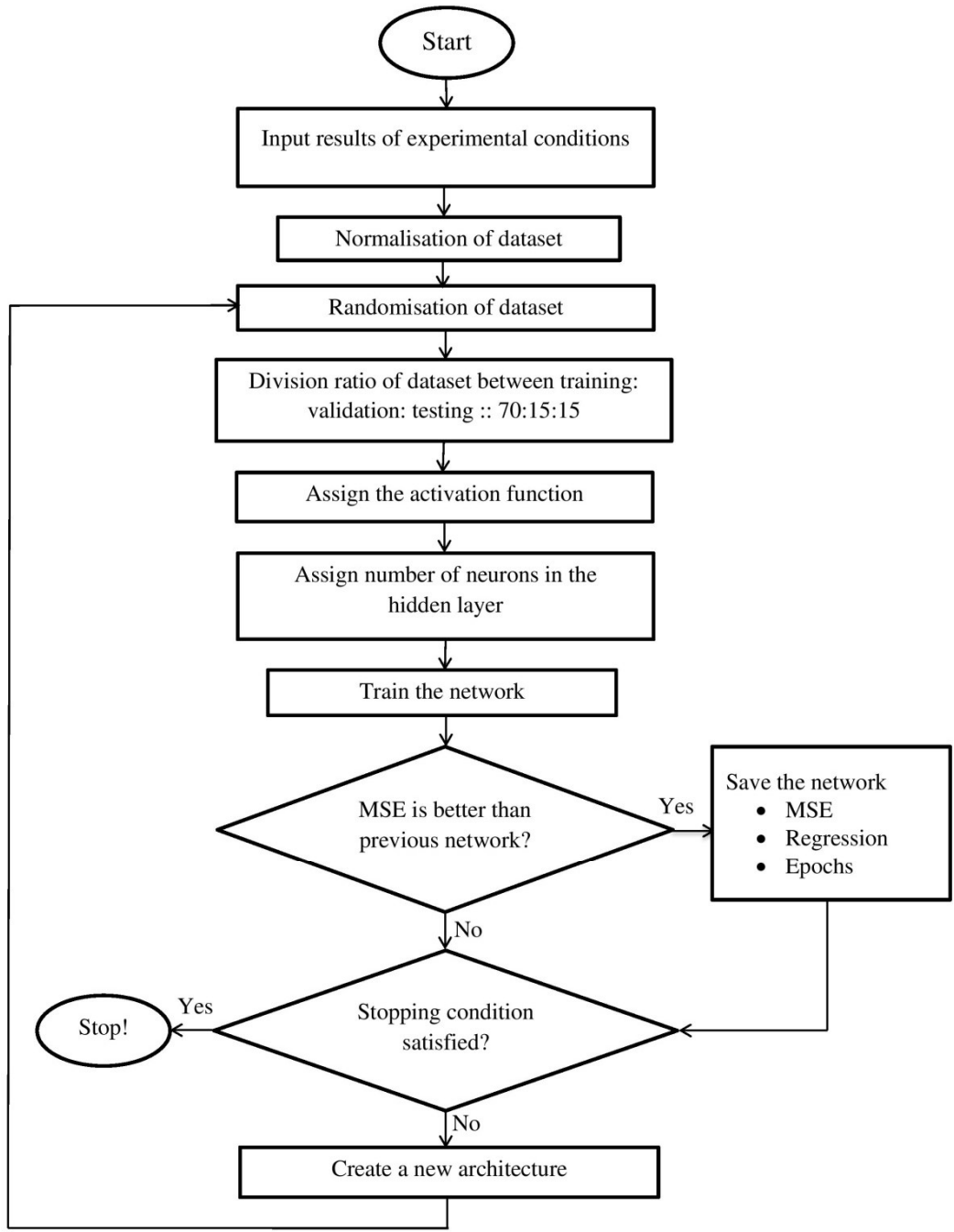

Figure 2. Flowchart of the proposed methodology.

This study also provides a comprehensive and rigorous approach on how to decide the optimum hidden layers and number of neurons in ANN based models and outlines a systematic method of choosing the optimum ANN architecture. Figure 2 represents the schematic flowchart of the proposed methodology. An optimum neural network architecture is proposed by varying the number of hidden layers, transfer functions and number of neurons in each hidden layer. Each ANN configuration has been trained with 100 independent runs to find the lowest training error, in order to minimize the chance of getting stuck in local minima in the ANN weight/bias term tuning process. The performance of the model can be evaluated by different accuracy measures such as the mean absolute error (MAE), root mean squared error (RMSE), normalised root mean squared error (NMSE) etc. However, each method has its own advantages and disadvantages. This aspect was very well explained by Azadi and Karimi-Jashni (2016). All of these quantitative measures summarises the error incurred in training and testing in a similar way. For training ANNs, MSE is the most popular choice of performance indicator and has been widely used in a wide variety of pattern recognition and machine learning problems (Bishop, 1995). The predictive accuracy of the model is evaluated by the MSE metric as given in equation (8).

$$MSE = \left( \sum_{i=1}^{n} \frac{(y_p - y_o)^2}{n} \right) \quad (8)$$

where, $n$ is number of datasets used for training the network, $y_p$ is mean of the predicted value and $y_o$ is the experimental (target) value.

Simulations were performed on a desktop workstation which consists of Intel i7- 3770 CPU, 3.4GHz processors with 4 GB of RAM. Parallelised simulation technique was used to optimise the capabilities of computer clusters (4 cores) using the parallel for loop.

Computational times for single and double layer model were also calculated and were approximately 18 hours and 190 hours ≈ 8 days respectively for 100 independent runs with re-shuffling the training datasets (100 times) and hidden layer with 30 neurons in the case of single layer and 15 neurons in each hidden layer for the double layer model. It is imperative to stress that the computational time reported here is the simulation time for finding the best model which train and cross validate multiple models with different number of layers and architecture to search for the best possible one.

Deciding the optimum ANN architecture is often tricky as there is always a chance of picking up inconsistent patterns and also a risk of premature convergence during the optimisation of the weight and bias terms of the FFNN. Therefore multiple randomisation of the optimiser with different initial guess and multiple shuffles of the data segmentation in training, validation, testing sets have been adopted here to enable higher accuracy and error estimates, in multiple Monte Carlo runs, to decide the best ANN architecture including the number of layers, neurons in each layer and the activation function, as also explored in (Das et al., 2012; Saha et al., 2012).

## 3. Results

### 3.1 Single layer MISO and MIMO models

The number of input and output parameter are nine and one respectively for the MISO model where the number of neurons in the hidden layer varies from 1 to 30 and LM algorithm is applied to train the neural networks. The models were trained and tested using both *tansig* and *logsig* nonlinear transfer functions in the hidden layer and *purelin* in the output layer. The dataset used to develop the ANN model contains 67 input/output patterns, out of which 70% (47 datasets) are used for training, 30% for testing and validation (10 datasets each) of the ANN model. Also, the input datasets are randomised for each and every iteration (100 independent runs were carried out). Simulations were also performed by varying the number of hidden layers in the model and transfer function. The networks are trained with varying number of hidden neurons in a hidden layer with different combinations of transfer functions. The performance of the network is evaluated on the basis of the MSE. The ANN architecture with the lowest MSE indicates a better model model (the best model is represented in the figures below by an arrow) in terms of predictive accuracy

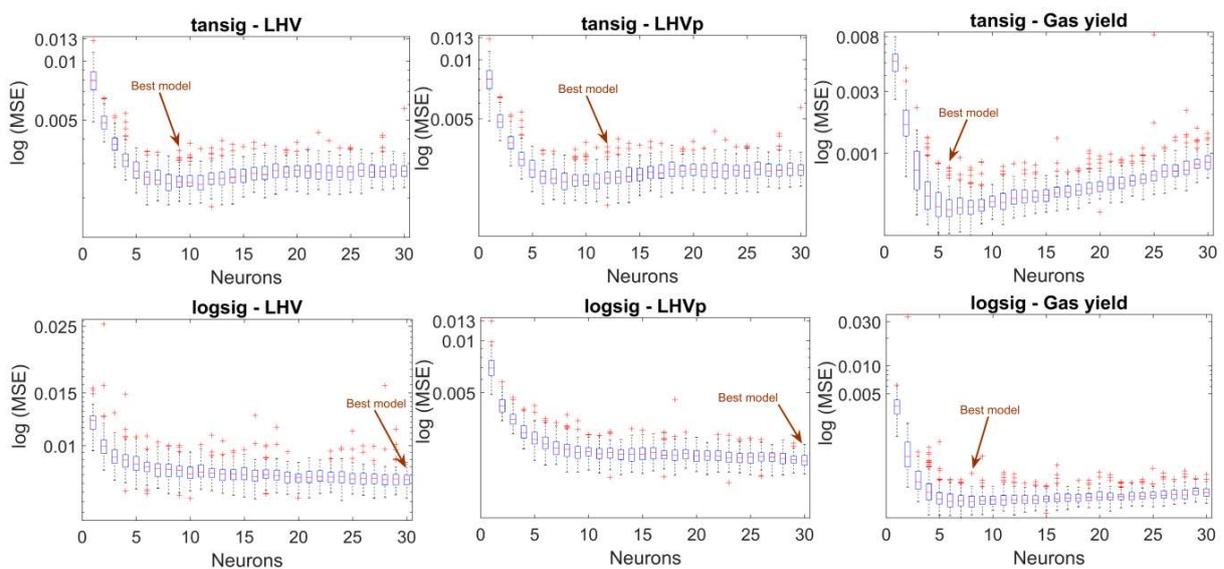

Figure 3. Box plot of single layer MISO model. Each box plot shows variation across multiple optimisation runs.

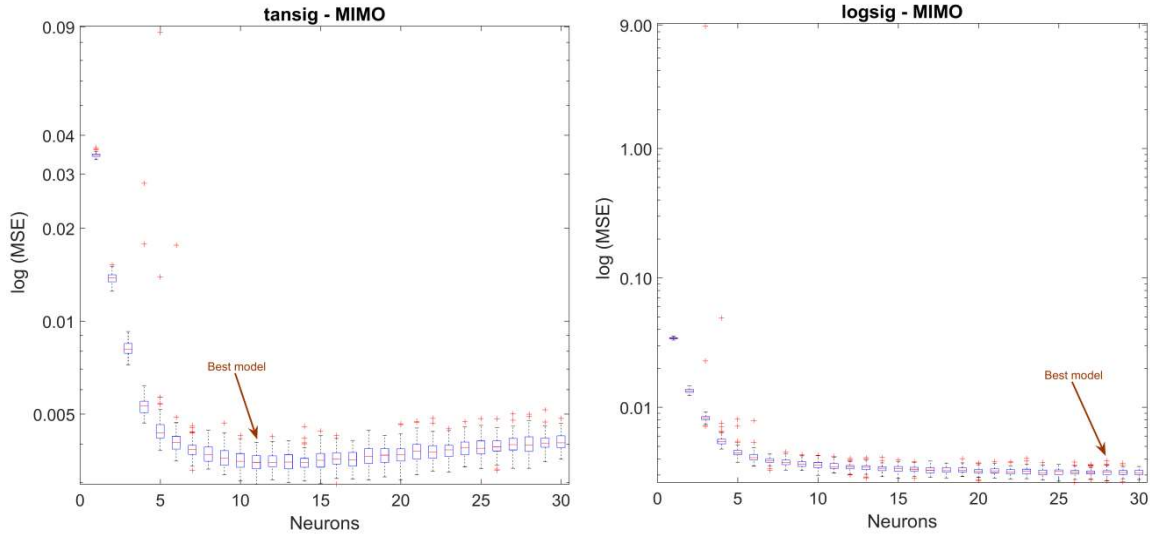

Figure 4. Box plot of single layer MIMO model. Each box plot shows variation across multiple optimisation runs.

The MSE of a single layer NN model is illustrated by box plots, as shown in Figure 3 & 4. Considering the variation of whiskers at different neuron numbers, a logarithmic scale is used on the ordinate axis for better representation of the graph. Box plots are used to display the distribution of data by minimum, first quartile, median, third quartile and maximum values. The central box comprises values between 25 and 75 percentiles and the whisker shows the range of values that fall within the maximum of 1.5 interquartile ranges (IQR). The band inside the box represents the median. The box plots often display the whole range of data starts from minimum to maximum, median and IQR. Box plots also display the outliers.

It can be seen from Figure 3 & 4 that increasing the number of neurons in the hidden layer does not imply that the model will have a better predictive accuracy (in the sense of the median across multiple optimisation runs). The best architecture for the neural network model is identified as that which has a minimum MSE (in terms of median of MSE). The minimum MSE with model details are presented in Table 2. It is evident from Table 2 that a single layer model with *logsig* transfer function has better accuracy compared to the *tansig* transfer function when used in the hidden layer. However, it is imperative to stress that models obtained using *tansig* are simpler than those obtained using *logsig* (i.e. the number of neurons are lower for the best model). Further simulations were performed with minimum MSE.

### 3.2 Double layer MISO and MIMO model

Figure 5 & 6 show the surface plots of the double layered MISO and MIMO models. Different combinations of transfer functions are used to find the best model. The minimum MSE with different combination of double layer NN models predicting the performances of the gasifier are presented in Table 2. As explained in Section 3.1, the best architectures are identified based on the lowest MSE for the subsequent simulations. Double layer MISO models show better predictive accuracy when the *logsig* transfer function was used in both the layers. Moreover, the MIMO model with the *tansig/logsig* (8/15 neurons in respective layers) combination has shown slightly better predictive accuracy.

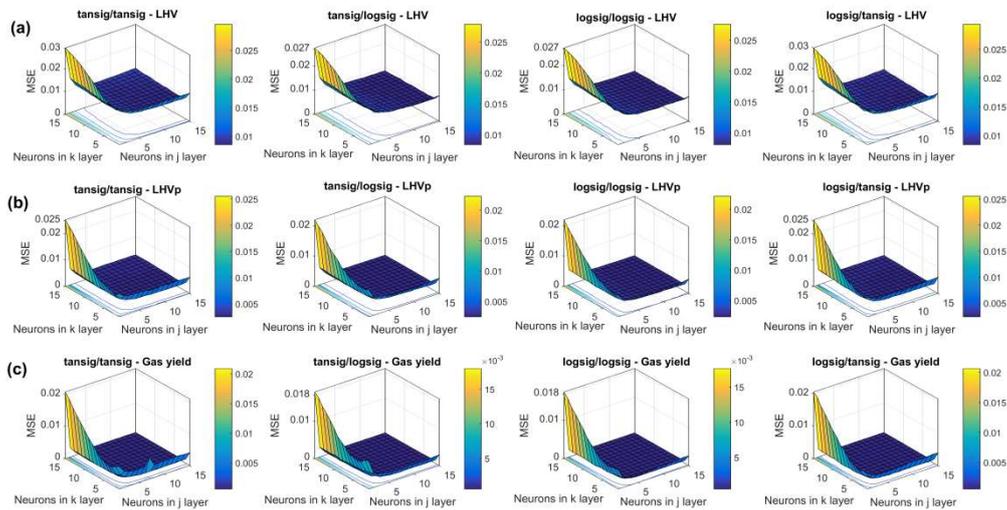

Figure 5. Surface plot of MSE for double layer MISO models (a) LHV (b) LHV$_p$ and (c) Gas yield.

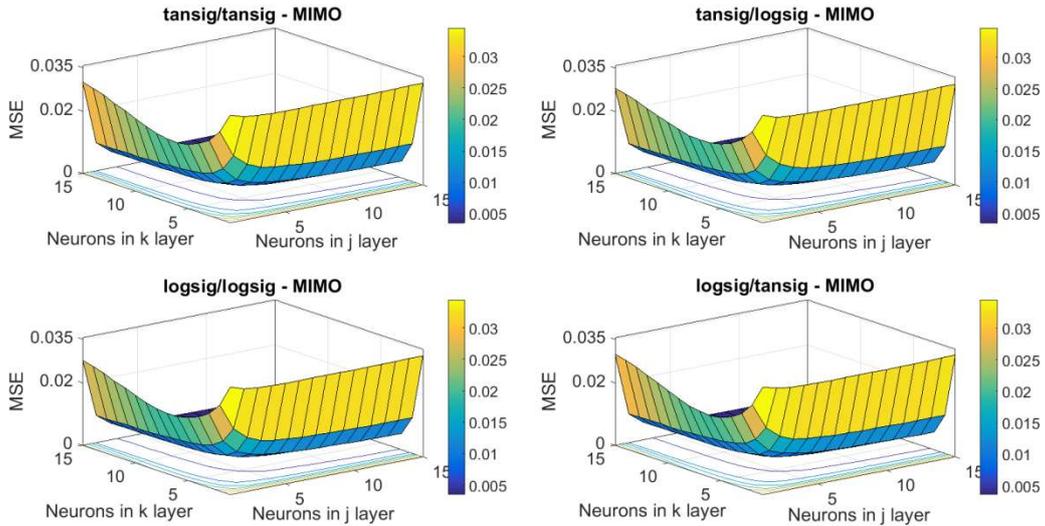

Figure 6. Surface plot of MSE for double layer MIMO models (LHV-LHVp-Gas yield).

Surface plots in Figure 5 & 6 show a three-dimensional view of the best prediction accuracy for the double layer model across a combination of different number of neurons in the respective layers. These plots are useful in finding the optimum combinations when an ANN regression model is fitted. It is used for the graphical visualisation of the smoothness of the fitted surface as the numbers of neurons in the layers are varied. The colour of the surface determined by the MSE is presented on the Z-axis. Contour maps of the MSE surface are presented below the surface plots to get a 2D visualization of the change in the predictive accuracy for the double layer NN model. The optimum number of neurons in the hidden layers for both the MISO and MISO models are tabulated in Table 2 based on the minimum reported MSE. The remaining discussion of this paper is based on the optimum architecture reported in Table 2.

Table 2. Training performance of the best ANN configuration

| ANN model | Number of Layer | Predictive parameter | Activation function | Number of neurons | Minimum MSE |
|---|---|---|---|---|---|
| MISO | 1 | LHV | *tansig* | 9 | 0.0086 |
| | | | ***logsig*** | **30** | **0.0077** |
| | | $LHV_p$ | *tansig* | 12 | 0.0024 |
| | | | ***logsig*** | **30** | **0.0021** |
| | | Gas yield | *tansig* | 6 | 0.0004 |
| | | | ***logsig*** | **8** | **0.0003** |
| MIMO | 1 | LHV-$LHV_p$-Gas yield | *tansig* | 11 | 0.0035 |
| | | | ***logsig*** | **28** | **0.0031** |
| MISO | 2 | LHV | *tansig/tansig* | 4/12 | 0.00852 |
| | | | *tansig/logsig* | 10/15 | 0.00837 |
| | | | ***logsig/logsig**** | **9/15** | **0.00810** |
| | | | *logsig/tansig* | 4/12 | 0.00840 |
| | | $LHV_p$ | *tansig/tansig* | 4/13 | 0.00251 |
| | | | *tansig/logsig* | 4/14 | 0.00247 |
| | | | ***logsig/logsig**** | **4/13** | **0.00229** |
| | | | *logsig/tansig* | 4/10 | 0.00234 |
| | | Gas yield | *tansig/tansig* | 6/6 | 0.00057 |
| | | | *tansig/logsig* | 6/10 | 0.00055 |
| | | | ***logsig/logsig**** | **6/5** | **0.00051** |
| | | | *logsig/tansig* | 5/6 | 0.00056 |
| MIMO | 2 | LHV-$LHV_p$-Gas yield | *tansig/tansig* | 6/12 | 0.00353 |
| | | | ***tansig/logsig**** | **8/15** | **0.00346** |
| | | | *logsig/logsig* | 7/15 | 0.00347 |
| | | | *logsig/tansig* | 7/15 | 0.00357 |

\* corresponds to the optimum NN model for the prediction of gasifier performance.

4. Discussion

Figure 7 depicts the representative case of the convergence characteristic of the ANN model for the LHVp (MISO, 4–13, *logsig/logsig* and LM algorithm). It can be seen that the MSE of the validation curve decreases slightly after 7 iterations. The validation fitness was found to increase after iteration 7 while predicting the LHVp in this particular case, indicating that the model would not generalise well if trained beyond this point. The model was trained to achieve an MSE of 0.001 with the prescribed number of neurons in the hidden layer as identified from Table 2. The double layer model has a *logsig* transfer function in each layer which has 4 and 13 neurons respectively. A similar approach was used while predicting the performance of the other output parameters.

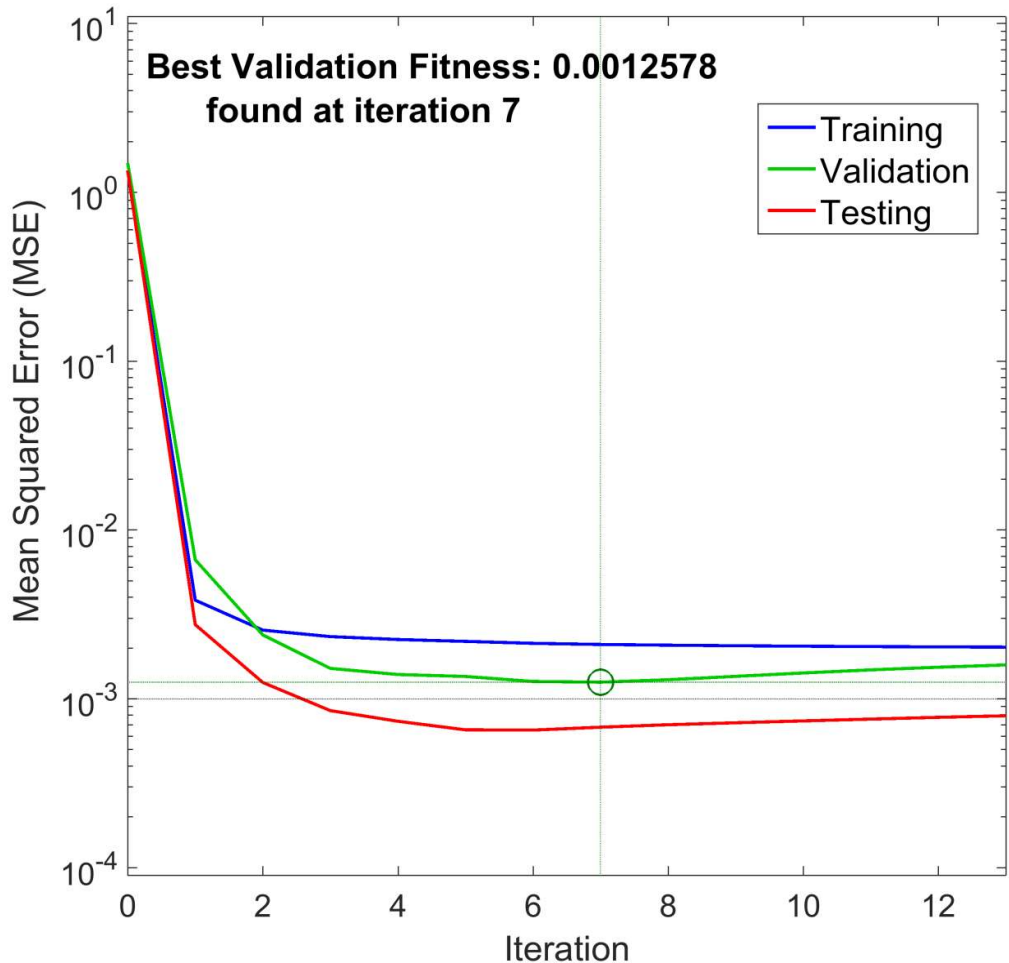

Figure 7. Convergence characteristics of the optimum double layer MISO model for the LHVp (4/13, logsig/logsig and LM algorithm).

## 4.1. Predictive performance of the single layer MISO ANN model

The optimum architecture for the MISO model for LHV, $LHV_p$ and gas yield is identified from Table 2 where it can be seen that a single layer model with *logsig* transfer function and LM learning algorithm has demonstrated better predictive accuracy. The optimum numbers of neurons in the hidden layers are 30, 30 and 8 for the LHV, $LHV_p$ and gas yield respectively.

The coefficient of determination ($R^2$) and MSE of training validation and testing datasets are reported in Figure 8 to predict the LHV, $LHV_p$ and syngas produced. Subplots show the experimental *vs.* ANN based model predicted values for the output calculation of the gas generated from the MSW gasification process. The $R^2$ value measures the performance of the model in predicting the output parameters from the experimental datasets. The plots (Figure 8) show that the degree of agreement between the experimental and predicted values for the training; validation and testing datasets are quite good (~ 90% or more). It is evident that most of the data-points lie on the straight line which indicates good performance of the developed model. It is clearly apparent that the accuracy of the network on training data is better than testing data. During the training mode the network always alters the values of its input and output weights to get the best fitness whereas in the testing phase (generalisation or validation) the output shows the actual predictive performance of the trained model on unseen data without adjusting the weights.

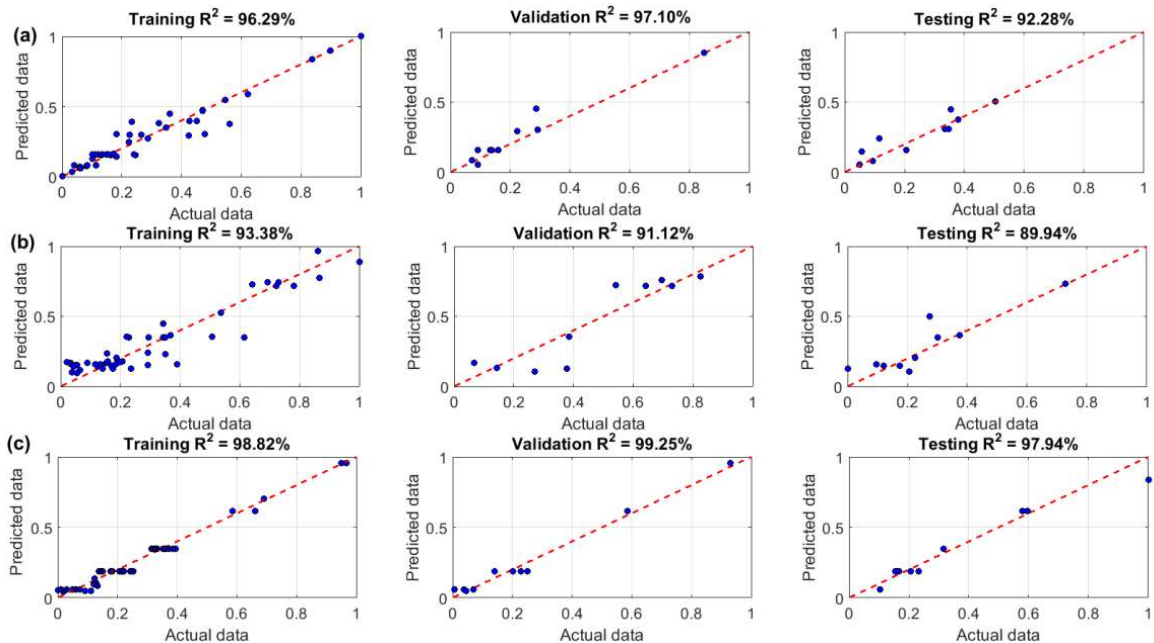

Figure 8. Prediction of single layer MISO ANN model with $R^2$ and MSE on the training, validation and the testing datasets (a) LHV (b) LHVp and (c) Gas yield.

## 4.2. Predictive capability of the single layer MIMO ANN model

This model is developed to predict multiple outputs by a single neural network. It can be seen from Table 2 that the *logsig* transfer function shows better accuracy compared with the *tansig* function. The actual *vs.* predicted output parameter from the best MIMO model on the training, validation and testing dataset have been reported in Figure 9. It shows the combined $R^2$ and MSE values. It can be observed that the generalisation and performance of the model is quite good. The evolved model has $R^2$ values over 94% in all three cases i.e. training, validation and testing. Although, a similar modelling paradigm is used while predicting the LHV, $LHV_p$ and syngas yield values separately, the MIMO model has a slightly better prediction accuracy ($R^2$ value over 98% on unseen data) on training, validation and testing over the MISO model. It is also evident that the MIMO model performed better compared to the three MISO models.

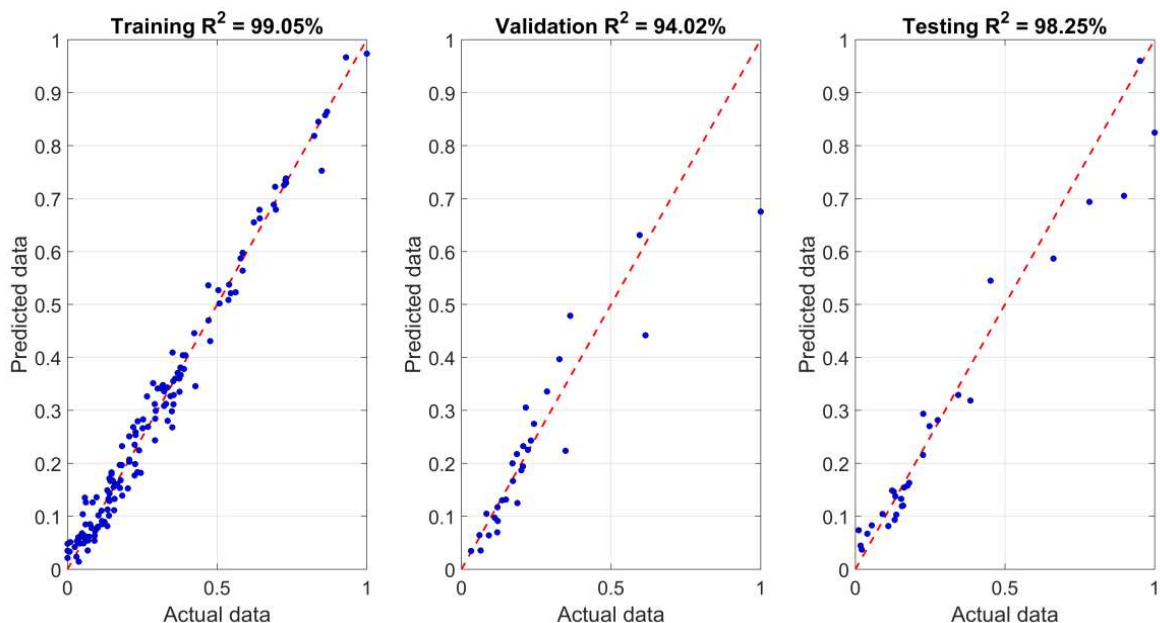

Figure 9. Prediction of single layer MIMO ANN model with $R^2$ and MSE on the training, validation and the testing datasets.

## 4.3. Predictive performance of the double layer MISO ANN model

Figure 10 shows the prediction performance of the trained double layered NN model reporting the actual *vs.* predicted values of LHV, LHVp and gas yield production. The simulations are performed at the best solution in Table 2, reporting minimum MSE and corresponding neural network architecture. It is noticed that the evolved models for $LHV_p$ and gas yield have slightly better predictive accuracy compared to LHV for the unseen datasets. The $R^2$ values for the testing and validation datasets are close to 100% for LHV and gas yield confirming the predictive reliability of the ANN model. In the case of the LHV prediction, $R^2$ for the validation dataset is found to be low, although the model generalised well over unseen datasets (testing) with an $R^2$ value of 96%. The plots in Figure 10 show that the degree of agreement between the experimental and predicted values for the training, validation and testing datasets are quite good with low relative error between the experimental and model predicted values under the cross-validation scheme.

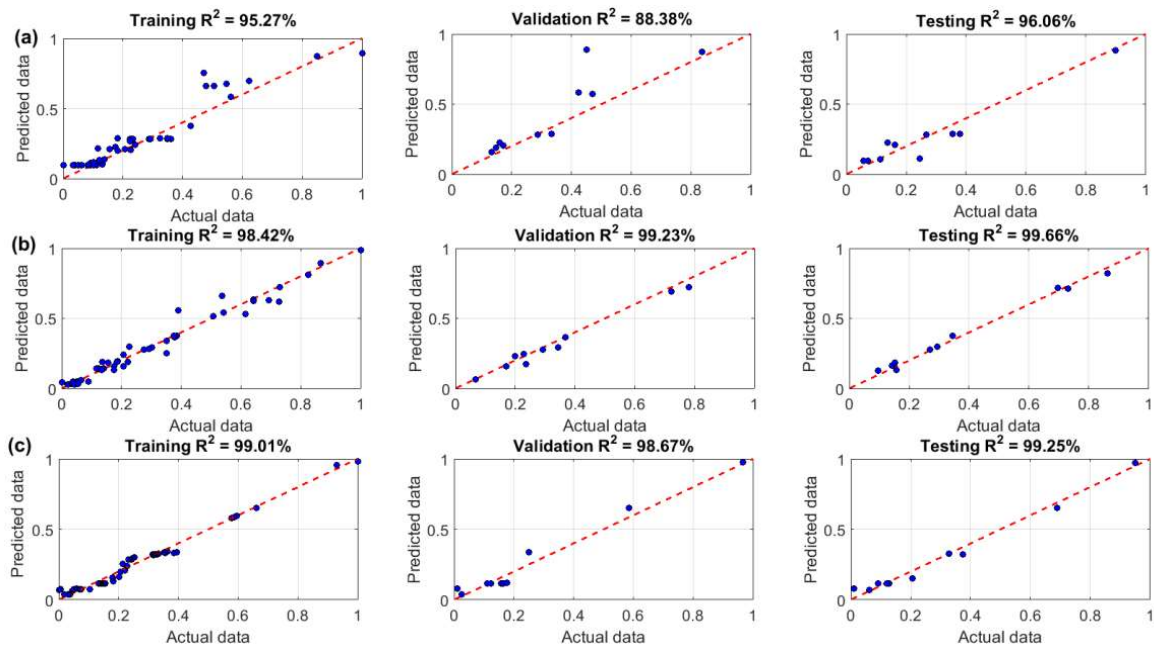

Figure 10. Prediction of double layer MISO based ANN model with $R^2$ and MSE on the training, validation and the testing datasets (a) LHV (b) LHVp and (c) Gas yield.

## 4.4. Predictive performance of the double layer MIMO ANN model

The training, validation and testing regression plots of the double layer MIMO model is illustrated in Figure 11. The trained MIMO model predicts the performance of the MSW gasification process using fuel characteristics and process parameters. The model used here contained 2 hidden layers consisting of 8 and 15 neurons in each layer with *tansig* and *logsig* as the activation functions in the first and second hidden layer respectively, which predicts the gasifier performance most accurately with respect to MSE criteria. The neural network was trained to predict three different output parameters (LHV, LHVp and gas yield). The degree of agreement ($R^2$ value) between experimental and simulated values justified the accuracy of the proposed ANN model.

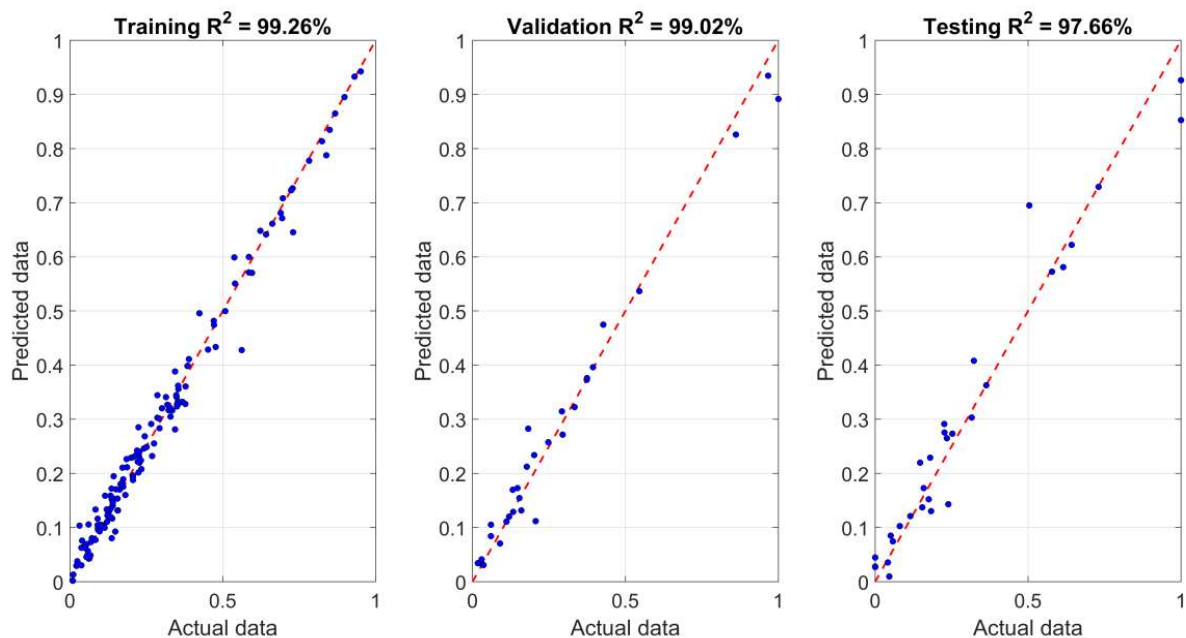

Figure 11. Prediction of double layer MIMO based ANN model with $R^2$ and MSE on the training, validation and the testing datasets.

## 4.5. Comparison of MISO and MIMO ANN models

The overall $R^2$ values for the different optimised structures and their corresponding MSE are reported in Table 3. It can be seen from Table 3 that the MIMO models show an improved performance over the MISO models. Although, the degree of agreement ($R^2$) for gas yield in the case of a single layer and the LHVp and gas yield for double layer models are higher compared to their respective MIMO model. Moreover, in both cases the MSE for the MIMO model is lower compared to the MISO model which measures the comparative performance of the two trained ANN modelling philosophies. Most of the studies reported to date (Puig-Arnavat et al., 2013; Xiao et al., 2009), focused on multiple input and single output. The comparative statistical analysis presented in Table 3 shows that the ANN model with multiple outputs has better prediction accuracy. It turns out that a single layer MIMO model with 28 neurons, LM training algorithm and *logsig* transfer function has better predictive accuracy amongst all four type of models explored, MISO, MIMO, single and double layer NNs. The computation time for finding the best model was about 18 hrs, whereas the double layer model took almost 200 hrs ≈ 8 days for the same simulation with the above reported computing hardware.

Table 3. Statistics of the best solutions of single and double layer model variants

| Number of layer | Algorithm $^\phi$ | Overall $R^2$ | MSE |
|---|---|---|---|
| 1 | 30/*logsig*/LHV | 95.95 | 0.00372 |
| | 30/*logsig*/LHV$_p$ | 92.56 | 0.00496 |
| | 8/*logsig*/Gas yield | 98.68 | 0.00109 |
| | 28/*logsig*/ MIMO | 98.05 | 0.00074 |
| 2 | 9/*logsig*/15/*logsig*/ LHV | 93.56 | 0.00157 |
| | 4/*logsig*/13/*logsig*/LHV$_p$ | 98.66 | 0.00203 |
| | 6/*logsig*/5/*logsig*/Gas yield | 98.95 | 0.00093 |
| | 8/*tansig*/15/*logsig*/MIMO | 98.90 | 0.00077 |

$^\phi$Number of neurons/ activation function/output parameter.

Despite the fact that ANN based models have advantages over traditional statistical approaches and have been widely used for similar prediction problems, they also have their own limitations. ANN based models are often referred to as black box models which are not capable of identifying the relative significance of the various parameters involved in the regression i.e. which input parameter influences the output most. The knowledge acquired during training of the model is intrinsic in nature and therefore it is difficult to draw a reasonable interpretation of the overall structure of the network. Furthermore, it also suffers from a greater computational burden, proneness to overfitting, and the empirical nature of model development (Tu, 1996).

5. **Conclusion**

In this study, MISO and MIMO ANN models, trained with the Levenberg-Marquardt back propagation algorithm are used to predict the LHV, $LHV_p$ and syngas yield from MSW in a fluidized bed gasifier using process parameters and elemental composition. It is shown that the predictive performance of the ANN models explored have a good agreement with the experimental datasets. This indicates that ANN can be used as an alternative method for modelling complex thermochemical processes. Good accuracy and performance of the trained ANN models (with $R^2 \approx 98\%$ for single layer and $R^2 \approx 99\%$ for double layer) have been achieved in all cases and the MSE is also found to be sufficiently low. The model has been tested against data from an atmospheric fluidized bed gasifier. The first application of this new approach has given a useful insight for equilibrium modelling however, calibration of the ANN model with more data is recommended since it is a self-adaptive, data-driven method with a few or no prior assumptions about the model structure. A simulation result for the presented study is quite promising and can be employed in learning and prediction of nonlinear complex mapping of gasification yields. This simulation paradigm illustrates the advantage of the proposed ANN model and can be exploited to simulate complex thermochemical processes such as gasification, pyrolysis and combustion.

The trained ANN model can be used for predicting the performance of similar kinds of gasifier operating under similar experimental condition. However, if the physical parameters in the input to the regression problem changes, the model needs to be retrained. Also, caution should be taken while developing the same ANN prediction model for heterogeneous data that comes partly or completely from different experimental protocols, which might need the breaking of the prediction problem into several smaller sub-problems that share some commonality between them, to improve prediction accuracy.

**Abbreviations:**

| | |
|---|---|
| LHV | Lower heating value |
| $LHV_p$ | Lower heating value of product gas including tars and entrained char |
| MSW | Municipal solid waste |
| ANN | Artificial neural network |
| LM | Levenberg–Marquardt |
| WtE | Waste to energy |
| FFNN | Feed forward neural network |
| CFD | Computational fluid-dynamics |
| SCG | Scaled conjugate gradient |
| BFGS | Broyden-Fletcher-Goldfarb-Shanno quasi-Newton |
| GDX | Gradient descent with momentum and adaptive learning rate |
| MIMO | Multiple input and multiple output |
| MISO | Multiple input and single output |
| tansig | Hyperbolic tangent sigmoid function |
| logsig | Logarithmic sigmoid function |
| purelin | Pure linear function |
| MSE | Mean squared error |
| IQR | Interquartile range |

**Acknowledgements**

The reported research has received funding from the People Programme (Marie Curie Actions) of the European Union's Seventh Framework Programme FP7/2007-2013/under REA grant agreement nº [289887]. The first author also acknowledges postgraduate research scholarship received from the University of Limerick, Ireland.